\documentclass[10pt,twocolumn,letterpaper]{article}

\usepackage[pagenumbers]{cvpr} 

\usepackage{siunitx,booktabs,graphicx,array}
\usepackage[table]{xcolor}

\definecolor{DeltaGreen}{HTML}{128C7E}

\newlength{\numcolwidth}
\newcolumntype{\numcol}{S[
  table-format=2.2,
  table-number-alignment=center,
  table-column-width=\numcolwidth
]}

\definecolor{cvprblue}{rgb}{0.21,0.49,0.74}
\usepackage[pagebackref,breaklinks,colorlinks,allcolors=cvprblue]{hyperref}

\usepackage[accsupp]{axessibility}
\usepackage{xcolor}  

\definecolor{robotaction}{RGB}{255, 140, 0}  
\definecolor{robottype}{RGB}{178, 34, 34}
\definecolor{robottask}{RGB}{65, 105, 225}  
\definecolor{controltype}{RGB}{0, 205, 0}
\definecolor{predsteps}{RGB}{216, 191, 216}
\definecolor{lavendermist}{rgb}{0.9, 0.9, 0.98}
\definecolor{visualtrace}{RGB}{255, 216, 0} 
\usepackage{colortbl}
\usepackage{booktabs}
\usepackage{color, soul}
\usepackage{graphicx}
\usepackage{amsmath}
\usepackage{empheq}
\usepackage{epigraph}
\usepackage{svg}
\usepackage{placeins}
\usepackage{float}
\definecolor{lightgray}{gray}{0.9}
\definecolor{lightblue}{rgb}{0.93,0.95,1.0}
\definecolor{darkgreen}{rgb}{0.0,0.6,0.0}
\definecolor{darkblue}{rgb}{0.0,0.0,0.5}
\definecolor{pinegreen}{rgb}{0.0, 0.47, 0.44}
\definecolor{deepmagenta}{rgb}{0.8, 0.0, 0.8}
\definecolor{amber}{rgb}{1.0, 0.49, 0.0}

\newcommand{\ignorebig}[1]{}

\def\Secref#1{Section~\ref{#1}}

\newcommand{\minisection}[1]{\noindent{\textbf{#1}.}}

\newlength\savewidth

\definecolor{lightred}{RGB}{241,140,142}
\definecolor{amber(sae/ece)}{rgb}{1.0, 0.49, 0.0}
\definecolor{battleshipgrey}{rgb}{0.52, 0.52, 0.51}
\definecolor{cadmiumorange}{rgb}{0.93, 0.53, 0.18}
\definecolor{applegreen}{rgb}{0.55, 0.71, 0.0}
\definecolor{cadmiumgreen}{rgb}{0.0, 0.42, 0.24}
\definecolor{forestgreen}{rgb}{0.13, 0.55, 0.13}
\definecolor{red}{rgb}{0.89, 0.0, 0.13}

\def\paperID{18413} 
\def\confName{CVPR}
\def\confYear{2026}

\title{Latent Implicit Visual Reasoning}

\author{
Kelvin Li$^{1*}$ \quad
Chuyi Shang$^{1*}$ \quad
Leonid Karlinsky$^{2}$ \quad
Rogerio Feris$^{3}$ \quad
Trevor Darrell$^{1}$ \quad
Roei Herzig$^{1, 3}$ \\
$^1$University of California, Berkeley \quad 
$^2$ Xero \quad
$^3$MIT-IBM Watson AI Lab \quad
}

\begin{document}
\maketitle
\begingroup
\renewcommand\thefootnote{}\footnotetext{$^*$Equal contribution}\addtocounter{footnote}{-1}
\endgroup
\begin{abstract}
While Large Multimodal Models (LMMs) have made significant progress, they remain largely text-centric, relying on language as their core reasoning modality. As a result, they are limited in their ability to handle reasoning tasks that are predominantly visual. Recent approaches have sought to address this by supervising intermediate visual steps with helper images, depth maps, or image crops. However, these strategies impose restrictive priors on what ``useful'' visual abstractions look like, add heavy annotation costs, and struggle to generalize across tasks. To address this critical limitation, we propose \textbf{Latent Implicit Visual Reasoning (LIVR)}, a task-agnostic mechanism that trains LMMs to discover and use \textbf{latent visual reasoning tokens} without explicit intermediate supervision. These tokens attend globally and re-encode the image in a task-adaptive way, enabling the model to extract relevant visual information without hand-crafted supervision. LIVR consistently outperforms direct supervised fine-tuning across diverse vision-centric tasks and multiple LMM backbones. In broader comparisons, LIVR remains competitive with or outperforms prior text-based and explicit-visual-intermediate reasoning methods, while requiring no additional intermediate supervision such as helper images, bounding boxes, image crops, depth maps, or chain-of-thought annotations. Our project page can be found here: https://www.chuyishang.com/livr/
\end{abstract}    
\begin{figure*}[t!]
\centering
\includegraphics[width=0.77\linewidth]{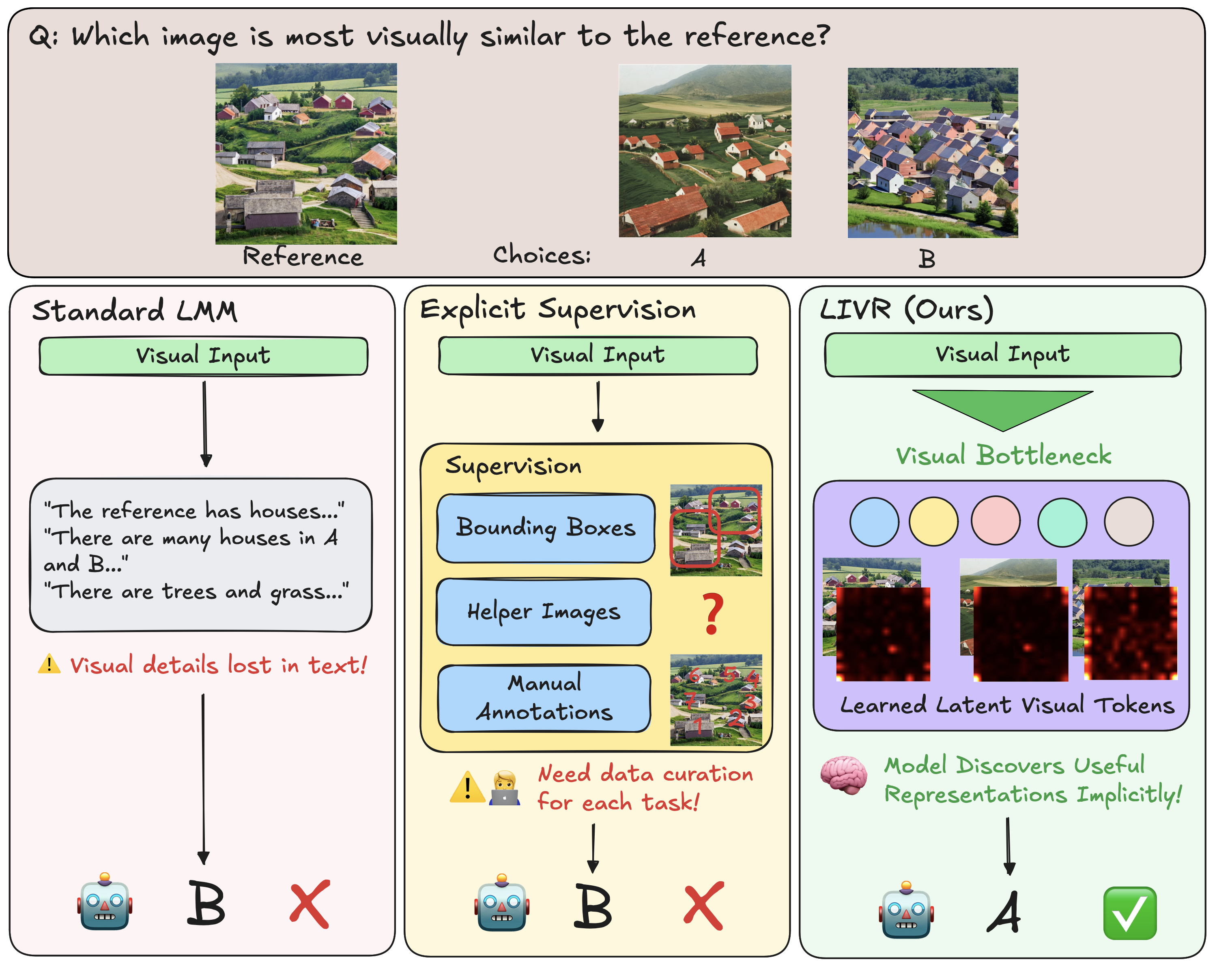}
  \caption{The model is asked to determine which image option is most similar to the reference image. Standard LMMs can only output text, which cannot capture all visual information and may introduce ambiguity. While methods using explicit supervision can train models to output intermediate reasoning steps, these approaches may fail when the reasoning steps themselves are unclear. Our approach allows the model to learn useful representations implicitly. Visualizing the attention maps of the latent tokens shows that the model has learned to recognize underlying visual structures relevant to answering the question that would have been hard for humans to design supervision for.}
    \label{fig:teaser}
\end{figure*}

\vspace{-10pt}

\section{Introduction}
\label{sec:intro}

In recent years, Large Multimodal Models (LMMs) have demonstrated great progress in visual understanding. However, they still struggle with vision-centric tasks that require fine-grained visual reasoning and abstraction. This limitation stems from several factors. First, most modern LMMs follow a LLaVA \cite{llava} style architecture, where visual inputs are projected into a language model that is trained to output text only. This introduces significant language bias, forcing the LMM to reason about visual information through text alone. Text-based representations may inherently lack the expressivity required to form the sophisticated visual abstractions needed for complex reasoning tasks. For example, humans can visualize objects from different angles, solve jigsaw puzzles, or identify visual patterns through mental imagery alone, without relying on language. Attempting to solve such tasks using language, may be extremely difficult. Moreover, recent LMM progress has largely focused on tasks requiring limited visual reasoning, such as document understanding or mathematical problem solving, where most of the reasoning occurs in the text space after initial visual information extraction.

Given these limitations, many works have attempted to train LMMs to be more ``visual'' through explicit intermediate supervision. However, this approach faces several challenges. First, it requires task-specific intermediate annotations, which can incur substantial annotation costs and limit scalability. Second, explicitly supervising intermediate steps embeds human assumptions about what constitutes ``useful'' visual reasoning. For example, models may be trained to predict bounding boxes, image crops, or other hand-designed visual targets, even though these representations may not be the most effective for the model to use. Third, for tasks requiring complex or abstract visual structure, it may be unclear even to humans what intermediate representation should be supervised. As a result, explicit-supervision approaches can be difficult to scale across diverse vision-centric tasks and may generalize poorly beyond the supervision regimes they were designed for.

Consider the task in Figure \ref{fig:teaser}, where the model is given a reference image and must select the most visually similar image from a set of choices. Describing the relationship between the sets of images using only text can be challenging and ambiguous. Training the model with explicit supervision is difficult as well since it is not clear what intermediate visual representations would be helpful to provide to the model. Even if we could identify useful intermediate steps, we would need to create large amounts of task-specific data, which is impractical to scale across different tasks.

Our proposed approach, Latent Implicit Visual Reasoning (LIVR), enables models to discover useful latent visual representations without hand-designed intermediate supervision. LIVR augments the LMM with latent tokens and trains them through a novel visual bottlenecking approach, which forces task-relevant visual information to pass through the latent tokens before the model predicts the answer. This allows the model to learn visual abstractions that improve downstream performance, without requiring humans to specify what the intermediate reasoning steps should be.

Our main contributions are as follows:
(i) We introduce LIVR, a new method for visual reasoning that allows the model to implicitly learn useful visual information through latent tokens, without the need for additional data or explicit supervision. (ii) In controlled data-matched experiments, we show that LIVR consistently outperforms direct supervised fine-tuning across nine perception-heavy tasks, three LMM backbones, and multi-task training. (iii) On external reasoning benchmarks, we show that LIVR remains competitive with or outperforms prior reasoning methods that rely on text-based reasoning or explicit visual intermediates, while avoiding costly intermediate supervision such as helper images, bounding boxes, image crops, or chain-of-thought annotations.
\vspace{-3pt}
\begin{figure*}[t!]
\centering
\includegraphics[width=\linewidth]{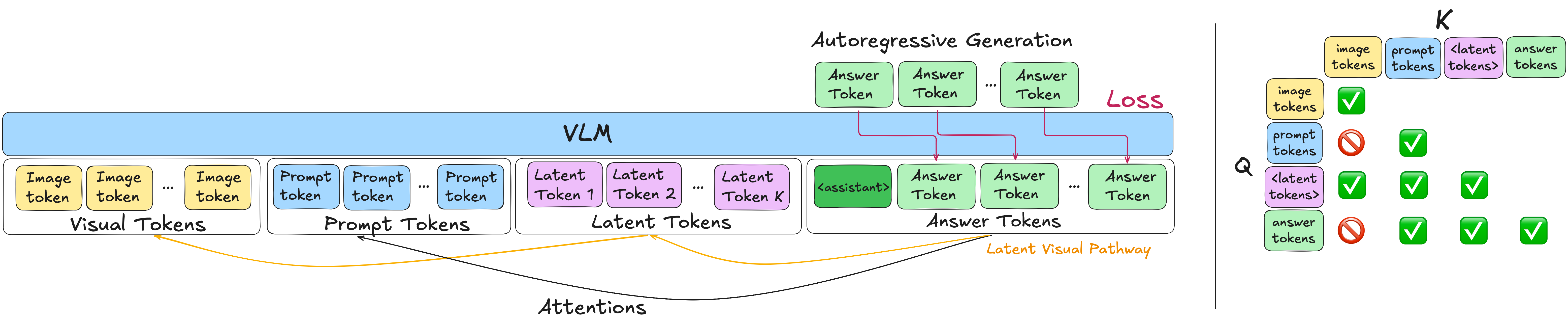}
  \caption{An illustration of our method and bottleneck attention masking. Latent tokens are appended to the prompt and losses are computed on the answer tokens. In our bottleneck attention masking, answers and prompt tokens cannot attend to image tokens.}
    \label{fig:method}
\end{figure*}

\section{Related Work}
\label{sec:related_work}

\noindent \textbf{Text-Based Visual Reasoning.}
Chain-of-thought (CoT) prompting has shown that explicitly generating intermediate text steps can substantially improve LLM performance on complex reasoning tasks \cite{cot, snell2024scalingllmtesttimecompute, NEURIPS2023_271db992}. Recent works extend CoT into the multimodal regime by training the model to describe its visual understanding in text before producing an answer \cite{llavacot, thawakar2025llamavo1rethinkingstepbystepvisual, yao2024mulberryempoweringmllmo1like}. For example, LLaVA-CoT fine-tunes an LMM to generate structured textual rationales that describe the image before concluding with an answer \cite{llavacot}. More recent works like Visual-RFT, Vision-R1, R1-VL and PAPO use RL-based post-training to encourage long, step-by-step textual explanations to help answer questions \cite{visual_rft,vision_r1,r1vl,wang2025perception}. In all of these approaches, the entire intermediate reasoning process is represented with text. Thus, it can be difficult for these methods to form rich, spatially structured visual abstractions that go beyond what can be easily verbalized.

\noindent \textbf{Interleaved Multimodal Reasoning.}
Text-only reasoning often struggles on visual tasks, motivating recent work that adds visual representations into the reasoning process. We group these methods into two main classes: region-based visual recycling and generated visual intermediates.

\noindent \textit{Region-Based Visual Recycling.} A common way to make multimodal reasoning more visual is to ground intermediate steps in image regions~\cite{visual_cot,argus,vgr,sarch2025vigorl,pixel-reasoner,uvcot,deepeyes}. These methods typically predict spatial coordinates, bounding boxes, or regions of interest, and then reuse the selected visual evidence during reasoning by cropping, zooming, or resampling parts of the original image. While this provides an intuitive mechanism for focusing visual computation, it fundamentally restricts the model to recycling information from localized regions of the original input. Moreover, it imposes a strong prior that useful visual reasoning should be expressible as one or more spatial regions. This prior is effective for tasks where the answer depends on a specific object or region of interest, but it is less natural for tasks requiring more global or abstract visual structure. In addition, when region annotations are required, such supervision can be costly to collect and difficult to scale across tasks.

\noindent \textit{Generated Visual Intermediates.} A second line of work goes beyond recycling existing image regions by generating new visual representations as intermediate reasoning steps. Some methods use multimodal generative backbones to explicitly generate intermediate images: MVoT~\cite{mvot} generates interleaved image visualizations during reasoning, while CoT-VLA~\cite{cotvla} generates future subgoal images before action prediction. Other methods operate inside LMMs with language-model backbones by injecting latent or discrete visual tokens. Mirage~\cite{mirage} and LVR~\cite{lvr} train latent visual tokens to reconstruct visual embeddings corresponding to a task-specific intermediate image or an image crop, while Aurora~\cite{aurora} adds discrete perception tokens that represent predefined visual quantities such as depth maps and bounding boxes. While these approaches expand reasoning beyond text, they still require specifying what the intermediate visual representation should be. This imposes a strong prior on the form of useful visual reasoning, whether as future frames, task-specific helper images, depth maps, or bounding boxes. Such supervision can be costly to obtain or derive, and may not exist for tasks where the useful intermediate is ambiguous or abstract. As a result, these methods can be difficult to generalize and scale across a wider range of vision-centric tasks.

\noindent \textbf{Latent Reasoning.}
A separate line of work explores allocating additional computation in latent space. Coconut treats hidden states as continuous ``thoughts'' that are iteratively fed back into the model \cite{coconut}, while Think Before You Speak uses pause tokens to trigger extra forward passes without emitting visible tokens \cite{pause_tokens}. Together, these works suggest that latent representations provide a more flexible internal representation space for reasoning than natural language, and that extra compute in latent space can be beneficial. Decoupling internal computation from external tokens lets the model refine its internal state solely to optimize task performance, rather than being constrained by what can be explicitly verbalized. Recent approaches have begun to explore latent-space reasoning in LMMs, but their latent variables are still trained with explicit intermediate supervision \cite{mirage, lvr}. In contrast, we study latent reasoning in an LMM without explicit supervision on intermediate solutions: dedicated latent tokens operate on joint visual--text states and are trained end-to-end from task objectives, allowing the model to learn implicit, task-specific visual abstractions.
\section{Method}
\label{sec:method}
We begin by describing some background on the LMM architectures (Section \ref{method:preliminaries}), then introduce our method (Section \ref{method:method}) and implementation details (Section \ref{method:implementation}). An illustration of our method is shown in Figure \ref{fig:method}.

\subsection{Preliminaries} \label{method:preliminaries}

\minisection{Large Multimodal Models} 
LMMs are generative models that process both visual and textual inputs to perform various tasks. They typically consist of three parts, a visual encoder, a language model decoder, and a projector that projects outputs from the visual encoder into the embedding space of the language model. To be more precise, given a text prompt $Q$ and visual input $I$, the prompt $Q$ is first encoded by a language encoder $l$. The image $I$ is encoded using a visual encoder $v$, then projected into the language model's embedding space via a projector $p$. Finally, the language model $M$ processes these embeddings to output a textual response $R$: 
\vspace{-2pt}
\[
R = M\left(p(v(I)),\, l(Q)\right)
\]

\minisection{Visual Question Answering}
We formulate each task as visual question answering. Given visual input $I$ and a text prompt $Q$, the model predicts an answer $a$, and performance is measured by whether the predicted answer matches the ground-truth answer.

\subsection{Latent Implicit Visual Reasoning}
\label{method:method}
Current LMMs are trained to autoregressively generate text tokens for visual tasks. Visual information is projected once into the language space at the beginning of the input, after which the LMM reasons primarily through text. We hypothesize that LMMs can benefit from additional visual computation space. However, rather than defining this space through explicit visual intermediates such as boxes, crops, depth maps, or helper images, we aim to let the model discover its own task-relevant visual representations. To do so, we equip the LMM with latent tokens and train the model to use them through a visual bottlenecking approach. This directly optimizes the latent tokens to capture whatever visual information is most useful for answering the question, without fixing what these representations should look like.

\minisection{Latent Tokens}
To provide the LMM with more expressivity to reason beyond the discrete text space, we equip it with latent tokens. Specifically, we introduce $K$ new special tokens, $L = \{l_1, l_2, \cdots, l_K\}$, to the model's existing vocabulary, $V$. The new vocabulary becomes $V \cup L$, with a total size of $|V| + K$. During training, we append these latent tokens to the input. Thus, given an original prompt $Q$, the new prompt $Q'$ becomes $Q \ + L$. While these tokens are randomly initialized, their corresponding rows in the embedding table remain unfrozen during training. Crucially, the model does not need to learn how to generate these latent tokens. Instead, it only needs to learn how to use them to represent important visual information.

\minisection{Visual Bottlenecking} \label{method:bottlenecking}
In order to train our latent tokens, we introduce a bottlenecking approach where we force visual information to pass through the latent tokens. We do this by modifying the attention mask so that the answer tokens can only attend to the prompt tokens $Q$ and the latent tokens $L$, but cannot attend to the visual inputs $I$. To avoid residual visual leakage to the answer tokens, we also prevent the prompt tokens $Q$ from attending to the visual inputs $I$. In this setup, the model can only ``see'' visual information through the latent tokens, which serve as the bottleneck. 

This bottlenecking is necessary because simply appending latent tokens does not guarantee that the model will use them; the model can still answer by attending directly to the original image tokens and may learn to ignore the added latents. By making the latent tokens the only pathway from the image to the answer, the model is forced to use the latent tokens, thereby making them useful.

\minisection{Multi-Stage Training}
We utilize a 2-stage approach to train our model. In Stage 1, we apply the masking described above and train the model using the standard negative log likelihood (NLL) objective:
\vspace{-2pt}
\[
\mathcal{L} = -\frac{1}{|x|} \sum_{i=1}^{|x|}\log P(x_i \mid x_{<i})
\]

where we compute the loss only on the answer tokens. By doing so, our objective directly optimizes the latent tokens to capture the \textit{most useful visual information for solving the question}.

After the latent tokens are trained to contain useful visual information in Stage 1, we revert to a standard attention mask that allows the answer tokens to attend to both the original image tokens and the latent tokens. The loss remains the same and is still computed only on the answer tokens. The goal of this stage is to train the model to jointly use both the original image tokens and the now-enriched latent tokens to answer the question.

\subsection{Implementation Details} 
\label{method:implementation}
We fine-tune the language backbone using LoRA (applied to attention and MLP blocks) \cite{hu2021loralowrankadaptationlarge, schulman2025lora}, while keeping the vision encoder and projector frozen. In addition to LoRA parameters, we unfreeze only the embedding rows corresponding to the $K$ new latent tokens. Full optimization and schedule details are provided in the Appendix.

\begin{table*}[!htbp]
\centering
\caption{Single-task fine-tuning accuracy.}
\label{tab:single_all_models}
\resizebox{\textwidth}{!}{
\begin{tabular}{l *{10}{\numcol}}
\toprule
Method & {Counting} & {Jigsaw} & {Local.} & {Vis. Corr.} & {Art Style} & {Sem. Corr.} & {Func. Corr.} & {Rel. Refl.} & {Vis. Sim.} & {Mean} \\
\midrule
Random Choice & 11.11 & 50.00 & 50.00 & 25.00 & 50.00 & 25.00 & 25.00 & 33.33 & 50.00 & 35.49 \\
\addlinespace[0.4em]

\multicolumn{11}{l}{\textbf{Qwen2.5\textnormal{-}VL\textnormal{-}3B\textnormal{-}Instruct}} \\
Zero-shot    & 46.78 & 49.33 & 56.56 & 29.86 & 55.56 & 32.37 & 26.71 & 45.52 & 50.37 & 43.67 \\
Direct SFT  & 60.04 & 53.33 & 75.41 & 88.00 & 83.76 & 41.01 & 18.49 & 44.78 & 89.63 & 61.61 \\
Ours        & \bfseries 63.64 & \bfseries 65.33 & \bfseries 79.51 & \bfseries 90.43 & \bfseries 87.18 & \bfseries 46.76 & \bfseries 31.51 & \bfseries 51.49 & \bfseries 94.82 & \bfseries 67.85 \\
$\Delta$ vs SFT
            & \multicolumn{1}{c}{\makebox[\numcolwidth][c]{\footnotesize\textcolor{DeltaGreen}{(+3.60)}}}
            & \multicolumn{1}{c}{\makebox[\numcolwidth][c]{\footnotesize\textcolor{DeltaGreen}{(+12.00)}}}
            & \multicolumn{1}{c}{\makebox[\numcolwidth][c]{\footnotesize\textcolor{DeltaGreen}{(+4.10)}}}
            & \multicolumn{1}{c}{\makebox[\numcolwidth][c]{\footnotesize\textcolor{DeltaGreen}{(+2.43)}}}
            & \multicolumn{1}{c}{\makebox[\numcolwidth][c]{\footnotesize\textcolor{DeltaGreen}{(+3.42)}}}
            & \multicolumn{1}{c}{\makebox[\numcolwidth][c]{\footnotesize\textcolor{DeltaGreen}{(+5.75)}}}
            & \multicolumn{1}{c}{\makebox[\numcolwidth][c]{\footnotesize\textcolor{DeltaGreen}{(+13.02)}}}
            & \multicolumn{1}{c}{\makebox[\numcolwidth][c]{\footnotesize\textcolor{DeltaGreen}{(+6.71)}}}
            & \multicolumn{1}{c}{\makebox[\numcolwidth][c]{\footnotesize\textcolor{DeltaGreen}{(+5.19)}}}
            & \multicolumn{1}{c}{\makebox[\numcolwidth][c]{\footnotesize\textcolor{DeltaGreen}{(+6.24)}}} \\
\addlinespace[0.6em]

\multicolumn{11}{l}{\textbf{Qwen3\textnormal{-}VL\textnormal{-}4B\textnormal{-}Instruct}} \\
Zero-shot    & 58.52 & 84.67 & 59.02 & 55.43 & 77.78 & 39.57 & 31.51 & 47.76 & 82.22 & 59.61 \\
Direct SFT  & 66.86 & 83.33 & 79.51 & 90.86 & 78.63 & 61.15 & 58.90 & 56.72 & 91.11 & 74.12 \\
Ours        & \bfseries 66.67 & \bfseries 85.33 & \bfseries 83.61 & \bfseries 93.29 & \bfseries 81.20 & \bfseries 64.75 & \bfseries 67.81 & \bfseries 62.69 & \bfseries 92.59 & \bfseries 77.55 \\
$\Delta$ vs SFT
            & \multicolumn{1}{c}{\makebox[\numcolwidth][c]{\footnotesize\textcolor{red}{(-0.19)}}}
            & \multicolumn{1}{c}{\makebox[\numcolwidth][c]{\footnotesize\textcolor{DeltaGreen}{(+2.00)}}}
            & \multicolumn{1}{c}{\makebox[\numcolwidth][c]{\footnotesize\textcolor{DeltaGreen}{(+4.10)}}}
            & \multicolumn{1}{c}{\makebox[\numcolwidth][c]{\footnotesize\textcolor{DeltaGreen}{(+2.43)}}}
            & \multicolumn{1}{c}{\makebox[\numcolwidth][c]{\footnotesize\textcolor{DeltaGreen}{(+2.57)}}}
            & \multicolumn{1}{c}{\makebox[\numcolwidth][c]{\footnotesize\textcolor{DeltaGreen}{(+3.60)}}}
            & \multicolumn{1}{c}{\makebox[\numcolwidth][c]{\footnotesize\textcolor{DeltaGreen}{(+8.91)}}}
            & \multicolumn{1}{c}{\makebox[\numcolwidth][c]{\footnotesize\textcolor{DeltaGreen}{(+5.97)}}}
            & \multicolumn{1}{c}{\makebox[\numcolwidth][c]{\footnotesize\textcolor{DeltaGreen}{(+1.48)}}}
            & \multicolumn{1}{c}{\makebox[\numcolwidth][c]{\footnotesize\textcolor{DeltaGreen}{(+3.43)}}} \\
\addlinespace[0.6em]

\multicolumn{11}{l}{\textbf{LLaVA\textnormal{-}OneVision\textnormal{-}1.5\textnormal{-}4B\textnormal{-}Instruct}} \\
Zero-shot    & 53.98 & 56.00 & 56.56 & 36.86 & 56.41 & 29.50 & 21.92 & 35.82 & 51.11 & 44.24 \\
Direct SFT  & 60.42 & 65.33 & 68.85 & 86.86 & 76.92 & 46.76 & 23.29 & 52.24 & 92.59 & 63.70 \\
Ours        & \bfseries 63.64 & \bfseries 70.67 & \bfseries 72.95 & \bfseries 88.71 & \bfseries 80.34 & \bfseries 51.08 & \bfseries 50.69 & \bfseries 53.73 & \bfseries 91.85 & \bfseries 69.30 \\
$\Delta$ vs SFT
            & \multicolumn{1}{c}{\makebox[\numcolwidth][c]{\footnotesize\textcolor{DeltaGreen}{(+3.22)}}}
            & \multicolumn{1}{c}{\makebox[\numcolwidth][c]{\footnotesize\textcolor{DeltaGreen}{(+5.34)}}}
            & \multicolumn{1}{c}{\makebox[\numcolwidth][c]{\footnotesize\textcolor{DeltaGreen}{(+4.10)}}}
            & \multicolumn{1}{c}{\makebox[\numcolwidth][c]{\footnotesize\textcolor{DeltaGreen}{(+1.85)}}}
            & \multicolumn{1}{c}{\makebox[\numcolwidth][c]{\footnotesize\textcolor{DeltaGreen}{(+3.42)}}}
            & \multicolumn{1}{c}{\makebox[\numcolwidth][c]{\footnotesize\textcolor{DeltaGreen}{(+4.32)}}}
            & \multicolumn{1}{c}{\makebox[\numcolwidth][c]{\footnotesize\textcolor{DeltaGreen}{(+27.40)}}}
            & \multicolumn{1}{c}{\makebox[\numcolwidth][c]{\footnotesize\textcolor{DeltaGreen}{(+1.49)}}}
            & \multicolumn{1}{c}{\makebox[\numcolwidth][c]{\footnotesize\textcolor{red}{(-0.74)}}}
            & \multicolumn{1}{c}{\makebox[\numcolwidth][c]{\footnotesize\textcolor{DeltaGreen}{(+5.60)}}} \\
\bottomrule
\end{tabular}}
\end{table*}

\section{Evaluation}
\label{sec:experiment}

We first evaluate LIVR in controlled single-task and multi-task settings using the same task data and backbones as Direct SFT. We then compare LIVR with prior reasoning methods on their respective benchmarks, followed by ablations and visualizations.

\subsection{Data-matched Experiments} 
\label{exp:experiments}

\minisection{Tasks and Datasets} \label{exp:datasets}
We evaluate our method on nine perception-heavy tasks adapted from the BLINK benchmark \cite{blink}: counting, jigsaw, object localization, visual correspondence, art style classification, semantic correspondence, functional correspondence, relative reflectance, and visual similarity. We choose these tasks because they require a strong degree of visual reasoning and abstraction. However, BLINK and most other challenging visual-centric datasets are designed for evaluation only, and there is a lack of readily available VQA-style training data. As such, we create our own training data sets from popular vision datasets. We note that all data we generate consists only of direct question-answer pairs, without any additional chains-of-thought or visual intermediate steps. All tasks except for counting are framed as BLINK-style multiple-choice VQA using top-1 accuracy as the evaluation metric; counting is evaluated in the standard open-ended setting with exact-match accuracy.

Counting uses the official PixMo-Count splits~\cite{pixmo}. We adopt PixMo-Count to evaluate a more challenging open-ended counting setting, where the model must generate the count rather than choose from discrete options. For the remaining tasks, we build training/validation splits from COCO~\cite{coco} (Jigsaw,
Localization), ArtBench-10~\cite{artbench} (art style), SPair-71k~\cite{spair71k}
(semantic correspondence), HPatches~\cite{hpatches} (visual correspondence),
FunKPoint~\cite{funkpoint} (functional correspondence), MID~\cite{mid} (relative
reflectance), and DreamSim~\cite{dreamsim} (visual similarity). We test on
the official BLINK validation sets for Jigsaw, Object Localization, Art Style,
Semantic Correspondence, Relative Reflectance, and Visual Similarity. For Visual
Correspondence and Functional Correspondence, we evaluate on held-out HPatches
and FunKPoint splits (rather than BLINK) due to the small size of these source
datasets. For all tasks, we de-duplicate custom train/validation data against
their corresponding test sets. Full construction details and prompt templates
are provided in the Appendix.

\minisection{Baselines and Models} \label{exp:baselines}
We experiment with three recent open-source LMMs of similar scale: Qwen2.5-VL-3B-Instruct \cite{qwen2.5vl}, Qwen3-VL-4B-Instruct \cite{qwen3technicalreport}, and LLaVA-OneVision-1.5-4B-Instruct \cite{llava_ov_1.5}. These models are competitive on a broad range of vision-language benchmarks, providing strong and comparable backbones for our study. For each task and backbone, we consider three settings: (i) \textbf{Zero-shot},
the pretrained instruct model evaluated without any task-specific training; (ii) \textbf{Direct SFT},
standard supervised fine-tuning on our task training set; and (iii) \textbf{LIVR}, our proposed
training method, run with the same task data and training setup as Direct SFT. Direct SFT serves as our main baseline, as it uses identical task supervision but no intermediate supervision, enabling a clean, data-matched comparison to LIVR.


\begin{table*}[t]
\centering
\caption{Multi-task fine-tuning accuracy on \textbf{Qwen3-VL-4B-Instruct}.}
\label{tab:multitask_q3}
\resizebox{0.75\textwidth}{!}{
\small
\begin{tabular}{l *{7}{\numcol}}
\toprule
Method & {Counting} & {Local.} & {Vis. Corr.} & {Sem. Corr.} & {Func. Corr.} & {Rel. Refl.} & {Mean} \\
\midrule
Zero-shot   & 58.52 & 59.02 & 55.43 & 39.57 & 31.51 & 47.76 & 48.64 \\
Direct SFT & 66.10 & 77.87 & 91.29 & 62.59 & 63.01 & 56.72 & 69.60 \\
Ours       & \bfseries 67.80 & \bfseries 81.97 & \bfseries 92.00 & \bfseries 67.63 & \bfseries 64.38 & \bfseries 60.45 & \bfseries 72.37 \\
$\Delta$ vs SFT
           & \multicolumn{1}{c}{\makebox[\numcolwidth][c]{\footnotesize\textcolor{DeltaGreen}{(+1.70)}}}
           & \multicolumn{1}{c}{\makebox[\numcolwidth][c]{\footnotesize\textcolor{DeltaGreen}{(+4.10)}}}
           & \multicolumn{1}{c}{\makebox[\numcolwidth][c]{\footnotesize\textcolor{DeltaGreen}{(+0.71)}}}
           & \multicolumn{1}{c}{\makebox[\numcolwidth][c]{\footnotesize\textcolor{DeltaGreen}{(+5.04)}}}
           & \multicolumn{1}{c}{\makebox[\numcolwidth][c]{\footnotesize\textcolor{DeltaGreen}{(+1.37)}}}
           & \multicolumn{1}{c}{\makebox[\numcolwidth][c]{\footnotesize\textcolor{DeltaGreen}{(+3.73)}}}
           & \multicolumn{1}{c}{\makebox[\numcolwidth][c]{\footnotesize\textcolor{DeltaGreen}{(+2.77)}}} \\
\bottomrule
\end{tabular}}
\end{table*}

\minisection{Single-Task Fine-Tuning} \label{exp:experiments:single_task}
For single-task experiments, we use 1k training examples per task. Direct supervised fine-tuning runs for 10 epochs. LIVR uses a two-stage schedule: 4 epochs of Stage 1 (visual bottlenecking) followed by 6 epochs of Stage 2 (standard masking) with $K=16$ latent tokens. These hyperparameters were determined through ablation studies on 3 tasks (Section \ref{exp:num_latents}) and kept fixed across all tasks, though we hypothesize that task-specific tuning could further improve results. For all runs, we select checkpoints by highest validation accuracy.

Table~\ref{tab:single_all_models} reports single-task accuracy across the nine visual-centric tasks for all three backbones. With Qwen2.5-VL, our method achieves significantly better results across all tasks, outperforming Direct SFT by an average of 6.24\%. The improvements are particularly pronounced on challenging tasks that require complex visual abstractions: gains of 12\% on Jigsaw and 13.02\% on Functional Correspondence demonstrate that our method effectively enhances the LMM's ability to form useful visual abstractions. We also observe gains on tasks such as Art Style, Visual Similarity, and Relative Reflectance, where explicit visual intermediates are difficult to specify; in these settings, LIVR provides a way to learn useful latent visual abstractions when it is hard—even for humans—to define hand-designed intermediate labels. On Qwen3-VL and LLaVA-OneVision-1.5, we also improve results across datasets by an average of 3.43\% and 5.60\% respectively, demonstrating the generalizability of our approach across multiple models.

\minisection{Multi-Task Fine-Tuning}\label{exp:experiments:multi_task}\label{sec:multitask} To test if our approach generalizes to multi-task setups, we use Qwen3-VL-4B-Instruct, the strongest backbone, and train on a combined dataset of six tasks: Counting, Localization, Visual Correspondence, Semantic Correspondence, Functional Correspondence, and Relative Reflectance, using 1k examples per task (6k total). We omit Jigsaw, Art Style, and Visual Similarity, as single-task baseline accuracies for Qwen3-VL-4B-Instruct on these tasks are already high, making relative improvements harder to interpret. Direct SFT is trained for 5 epochs, while LIVR is trained for 2 epochs of Stage~1 and 3 epochs of Stage~2, maintaining the same 2:3 ratio as in single-task experiments and using $K=16$ latent tokens. We report performance using the final checkpoint.

Table~\ref{tab:multitask_q3} shows results for multi-task training on Qwen3-VL-4B-Instruct across the six perception tasks. LIVR improves over Direct SFT on all tasks, demonstrating that the latent mechanism that is effective in single-task settings also benefits joint multi-task training. A key advantage of LIVR is its task-agnostic nature: because it trains latent tokens implicitly from the end-task loss without requiring task-specific helper images or intermediate labels, the same method applies directly to multi-task settings. This contrasts with approaches that tie latent tokens to task-specific visual targets (e.g., depth maps, bounding boxes, helper images), which require different supervision per task and are difficult to extend to heterogeneous multi-task setups. This makes our method well-suited as a simple, general-purpose enhancement for perception-heavy multi-task fine-tuning.

\subsection{Comparison with Prior Reasoning Methods}
\label{exp:prior_comparisons}

\minisection{Visual Spatial Planning}
We compare LIVR with Mirage \cite{mirage}, a latent reasoning approach that trains latents to represent intermediate helper images. We evaluate on the Visual Spatial Planning (VSP) task only, as data for other tasks have not been released. For LIVR, we discard the helper images and set the number of latents to match Mirage at $K=4$. On Qwen2.5-VL-3B, we reproduce Mirage using their released dataset and helper images; zero-shot accuracy is 6.00, Mirage achieves 46.00, and LIVR reaches 66.00 (+20.00). On Qwen2.5-VL-7B, we use Mirage's reported numbers; LIVR achieves 77.50, outperforming Mirage (76.00), Mirage with text-CoT (58.00), and Mirage with RL (60.00), despite not using any helper images.

\begin{table}[!htbp]
\vspace{-2mm}
\centering
\caption{Comparison across spatial reasoning benchmarks. All rows except LIVR-3B (Ours) are reported from ViGoRL.}
\label{tab:vigorl}
\vspace{-1mm}
\setlength{\tabcolsep}{4pt}
\renewcommand{\arraystretch}{0.95}
\footnotesize
\resizebox{\columnwidth}{!}{
\begin{tabular}{lccc}
\toprule
\textbf{Method} & \textbf{SAT Val} & \textbf{BLINK-3} & \textbf{RoboSpatial} \\
\midrule
Qwen2.5VL-3B                 & 46.1  & 44.4  & 54.4 \\
+ SFT direct                 & 58.3  & 46.4  & 62.3 \\
+ Vanilla GRPO               & 50.0  & 46.5  & \textbf{69.7} \\
Text-CoT (SFT+GRPO)          & 58.7 & 45.4 & --   \\
ViGoRL-3B                    & \underline{62.9}  & \underline{48.5}  & \underline{67.1} \\
\midrule
\addlinespace[1.5pt]
\textbf{LIVR-3B (Ours)}      & \textbf{85.6} & \textbf{59.5} & 66.7 \\
\bottomrule
\end{tabular}
}
\vspace{-2mm}
\end{table}
\begin{table}[!htbp]
\vspace{-2mm}
\centering
\caption{Comparison across MMVP, V*, and BLINK benchmarks. All rows except LIVR-7B (Ours) are reported from LVR.}
\label{tab:lvr_comparison}
\vspace{-1mm}
\setlength{\tabcolsep}{5pt}
\renewcommand{\arraystretch}{0.95}
\footnotesize
\begin{tabular*}{\columnwidth}{@{\extracolsep{\fill}}lccc@{}}
\toprule
\textbf{Method} & \textbf{MMVP} & \textbf{V*} & \textbf{BLINK-5} \\
\midrule
Qwen2.5-VL-7B        & 66.7 & 78.5 & 53.66 \\
PAPO                 & 54.3 & 36.1 & \underline{54.81} \\
Vision-R1            & 46.7 & 70.2 & 42.76 \\
PixelReasoner        & 67.0 & \underline{80.1} & 54.52 \\
LVR-7B               & \underline{71.7} & \textbf{80.6} & \textbf{55.37} \\
\specialrule{0.4pt}{2pt}{2pt}
\textbf{LIVR-7B (Ours)} & \textbf{75.3} & \underline{80.1} & 54.28 \\
\bottomrule
\end{tabular*}
\vspace{-3mm}
\end{table}
\minisection{Spatial Reasoning Generalization}
In Table \ref{tab:vigorl}, we compare LIVR against several baselines: direct SFT, vanilla GRPO \cite{deepseek-math}, text SFT+RL and ViGoRL \cite{sarch2025vigorl}, a method trained with SFT and RL to output textual reasoning and bounding box coordinates as intermediate steps. All methods are initialized from Qwen-2.5-VL-3B and trained on the same SAT-32k dataset \cite{ray2025satdynamicspatialaptitude}, taken from ViGoRL. Because SAT is fully synthetic, it serves as a useful test for out-of-distribution generalization to real images like in BLINK \cite{blink}. To isolate the gains from latent reasoning, we remove intermediate images and textual CoT from LIVR, while other methods are free to use them. Following ViGoRL, we evaluate on SAT Val, a 3-task subset of BLINK (relative depth, multi-view reasoning, spatial relation), and the configuration and compatibility splits of RoboSpatial \cite{song2025robospatial}, which tests spatial understanding in robotics contexts. Despite operating without text-CoT, explicit grounding or RL, LIVR achieves strong results across all benchmarks and demonstrates the ability to generalize to out-of-distribution tasks.

\minisection{Broader Visual Reasoning Benchmarks} Table \ref{tab:lvr_comparison} compares our method with LVR (latent visual intermediate + RL), PixelReasoner (visual intermediate + RL), Vision-R1 and PAPO (text CoT + RL). All use Qwen-2.5-VL-7B. We train LIVR on the Visual-CoT \cite{visual_cot} training set used by LVR, and removed intermediate images and text CoT. Following LVR, we evaluate on MMVP, V*, and 5 BLINK subsets (Counting, IQ Test, Jigsaw, Rel. Reflectance, Spatial Rel.). LIVR is competitive or better across all tasks despite not using costly intermediate supervision such as image crops, visual intermediates, or text CoT, demonstrating its generalizability and effectiveness.

\subsection{Ablations and Additional Experiments} \label{exp:ablations}
\subsubsection{Usefulness of Latent Tokens} 
\label{abl:latents_used}
We next test whether the model truly relies on latent tokens rather than ignoring them. We compare LIVR against a \emph{latents-only} variant that adds $K=16$ latent tokens but trains only with Stage~2 (no bottlenecking). This control is designed to match LIVR's added capacity while providing no explicit pressure for latent tokens to carry visual information, creating an ``unused-latents'' baseline. We report results on the Localization task using Qwen3-VL-4B-Instruct.

When latent tokens are removed at evaluation, the latents-only model maintains the same accuracy (79.51 $\rightarrow$ 79.51), indicating it has learned to ignore the extra tokens. In contrast, in the standard (unmasked) setting LIVR achieves higher accuracy than the latents-only model (83.61 vs.\ 79.51) and suffers a clear drop when latents are removed (83.61 $\rightarrow$ 76.23), showing that it depends on them. This is further confirmed by attention patterns: measuring the mean attention from answer tokens to latent tokens (averaged over all heads, layers, and positions), we find much higher scores for LIVR than for the latents-only model (0.076 vs.\ 0.028). To test whether latents encode useful visual information, we evaluate both models under a bottleneck mask at test time, where the model can only view the image through latent tokens. Under this bottleneck, the latents-only model performs on par with random guessing (43.44), indicating its latents carry no useful visual information, while LIVR retains much higher accuracy (70.49). As a sanity check, if we additionally drop latent tokens under the bottleneck mask, accuracy falls to 43.44, since the image pathway is removed entirely. Together, these results show that the latents in our method are both actually used by the model and encode task-relevant visual information.

\subsubsection{Design Ablations for LIVR}
\label{sec:design_ablations}
We individually test the effectiveness of the two main components of our approach, latent tokens and bottlenecking. We perform these ablations using Qwen3-VL-4B-Instruct as our base model across three challenging tasks: Localization, Semantic Correspondence, and Functional Correspondence. The results are displayed in Table~\ref{tab:design_ablations}.

\minisection{Bottleneck Ablation}
We first revisit the \emph{latents-only} variant described in Section~\ref{abl:latents_used}, which adds latent tokens but skips Stage~1 bottlenecking. This isolates the effect of added capacity. However, simply introducing extra tokens without bottleneck training significantly underperforms LIVR, showing that capacity alone is insufficient.

\minisection{Latent Ablation}
Second, we test a \emph{mask-only} variant that applies the Stage~1 bottleneck without adding latent tokens. Here, answer tokens cannot attend directly to vision tokens, but prompt tokens can still see the image. The goal is to force existing prompt tokens to act as a visual bottleneck without adding new capacity. This variant also underperforms LIVR. A plausible explanation is that existing text tokens already carry pre-trained semantics, making them harder to repurpose to form abstract visual representations. In contrast, newly introduced latent tokens are free to adapt and can more easily learn to form rich visual abstractions.

Together, these results suggest that both the dedicated latent tokens and the visual bottleneck are necessary for LIVR's full gains. For completeness, we include three additional controls in the same table: duplicating the input image tokens (``input image twice''), where we concatenate two copies of the same image tokens at both training (with and without 2-stage masking) and inference as a generic control for extra visual compute, and prompt tuning \cite{prompt_tuning}, a lightweight adaptation baseline.

\begin{table}[t]
\centering
\caption{Design ablations and additional controls.}
\label{tab:design_ablations}
\resizebox{\linewidth}{!}{
\begin{tabular}{l
                S[table-format=2.2]
                S[table-format=2.2]
                S[table-format=2.2]}
\toprule
Method & {Local.} & {Sem. Corr.} & {Func. Corr.} \\
\midrule
Baseline               & 59.02 & 39.57 & 31.51 \\
Direct SFT             & 79.51 & 61.15 & 58.90 \\
Ours                   & \bfseries 83.61 & \bfseries 64.75 & \bfseries 67.81 \\
Latents only (no mask) & 79.51 & 61.15 & 58.22 \\
Mask only (no latents) & 80.33 & 61.16 & 59.59 \\
Input image x2 (no mask)      & 78.69 & 61.16 & 58.22 \\
Input image x2 (2-stg mask) & 77.87 & 61.87 & 56.85 \\
Prompt tuning          & 71.31 & 49.64 & 36.30 \\
\bottomrule
\end{tabular}}
\end{table}

\begin{table*}[t]
\centering
\caption{\textbf{Ablations of latent-token design choices on Qwen3-VL-4B-Instruct.} All numbers are accuracies (\%).}
\label{tab:latent_ablations}
\renewcommand{\arraystretch}{1.15}
\small

\makebox[\textwidth][c]{\hspace*{-1cm}%
\begin{subtable}[t]{0.40\textwidth}
    \centering
    \caption{Masking Strategy}
    \label{tab:abl_masking}
    \begin{tabular}{@{}lccc@{}}
    \toprule
    Method & Loc. & Sem. & Func. \\
    \midrule
    Ans$\to$Vis only & 77.87 & 60.43 & 60.27 \\
    \rowcolor{gray!10}
    \textbf{Ans+Prompt$\to$Vis (ours)} & \textbf{83.61} & \textbf{64.75} & \textbf{67.81} \\
    Ours+Latent$\to$Prompt & 81.15 & 62.59 & 63.01 \\
    \bottomrule
    \end{tabular}
\end{subtable}%
\begin{subtable}[t]{0.3\textwidth}
    \centering
    \caption{Stage-1 / Stage-2 Epochs}
    \label{tab:abl_stages}
    \begin{tabular}{@{}cccc@{}}
    \toprule
    (S1, S2) & Loc. & Sem. & Func. \\
    \midrule
    0, 10 & 79.51 & 61.15 & 58.22 \\
    2, 8 & 80.33 & 58.27 & 65.75 \\
    \rowcolor{gray!10}
    \textbf{4, 6} & \textbf{83.61} & \textbf{64.75} & \textbf{67.81} \\
    6, 4 & 81.15 & 61.87 & 66.44 \\
    8, 2 & 77.87 & 59.71 & 60.27 \\
    \bottomrule
    \end{tabular}
\end{subtable}
\begin{subtable}[t]{0.2\textwidth}
    \centering
    \caption{Number of Latents}
    \label{tab:abl_numlatents}
    \begin{tabular}{@{}cccc@{}}
    \toprule
    \# Lat. & Loc. & Sem. & Func. \\
    \midrule
    4 & 81.15 & 62.59 & 66.40 \\
    8 & 80.33 & 63.31 & 67.12 \\
    \rowcolor{gray!10}
    \textbf{16} & \textbf{83.61} & \textbf{64.75} & \textbf{67.81} \\
    32 & 80.33 & 62.59 & 63.01 \\
    \bottomrule
    \end{tabular}
\end{subtable}%
}
\end{table*}

\subsubsection{Architectural and Training Choices}
\label{sec:latent_ablations_sec}

We ablate design choices of LIVR on Qwen3-VL-4B-Instruct, again focusing on Localization, Semantic Correspondence, and Functional Correspondence. We vary each design choice independently, keeping all others fixed to our defaults: latents placed after the prompt, our default masking scheme (blocking both answer-to-vision and prompt-to-vision attention), a 4-epoch Stage~1, 6-epoch Stage~2 schedule, unshared latent embeddings, and $K=16$ latent tokens.

\minisection{Position of latents}
We compare placing latents before versus after the prompt. Placing latents after the prompt (our default) as opposed to before the prompt yields higher accuracy across all three tasks. Specifically, for the Localization, Semantic Correspondence, and Functional Correspondence tasks, we get scores of (83.61 vs. 80.33), (64.75 vs. 61.87), and (67.81 vs. 63.70), respectively, for placing the latents after vs. before the prompts. We hypothesize that when latents appear before the prompt, they cannot condition on the question and are farther from answer tokens, making them harder for the model to exploit effectively.

\minisection{Masking strategy}
Table~\ref{tab:latent_ablations}(a) compares three masking schemes. Our default approach blocks both answer-to-vision and prompt-to-vision attention, forcing all visual information to flow through latents, and achieves the best performance. Blocking only answer-to-vision attention is insufficient: visual information can still reach answer tokens via the prompt, so latents never become a true bottleneck. Conversely, further blocking latents from attending to the prompt is too restrictive, as latents need to see the question to determine what visual information to encode.

\minisection{Stage-1 / Stage-2 schedule}
For our main experiments, we train the model for 4 epochs in Stage~1 and 6 epochs in Stage~2. We experiment with different allocations of Stage~1 and Stage~2 epochs in Table~\ref{tab:latent_ablations}(b), while keeping the total number of epochs at $10$.
Using only Stage~2 (0,10) corresponds to the latents-only setting from Section~\ref{abl:latents_used} and underperforms LIVR, again highlighting the importance of bottleneck training.
Conversely, an (8,2) split also hurts; we hypothesize that in this case the model does not have enough Stage~2 training to learn how to integrate the latent representations with the original image tokens under the standard mask.
A balanced schedule with $4$ Stage~1 and $6$ Stage~2 epochs provides the best trade-off, giving latents enough time to learn visual information while still allowing ample joint training with standard masking.

\minisection{Shared vs.\ unshared latent embeddings}
In our method, we use different embeddings for each of our $K$ latent tokens. However, we can also insert the same latent token $K$ times, in a configuration we call "shared embeddings". We find that using unshared embeddings (one learnable embedding per latent) yields higher accuracy compared to shared embeddings across all 3 tasks. Specifically, we have scores of (83.61 vs. 81.15), (64.75 vs. 61.87), and (67.81 vs. 63.70) for the Localization, Semantic Correspondence, and Functional Correspondence tasks, respectively. This is consistent with the idea that giving each latent its own embedding increases the expressivity of the latent set.

\minisection{Number of latents} \label{exp:num_latents}
For our standard experiments, we set $K=16$, inserting 16 latent tokens per prompt. We experiment with varying $K$ by using values of $4, 8, 16, 32$, which is displayed in Table~\ref{tab:latent_ablations}(c). Accuracy generally improves as $K$ increases from $4$ to $16$, with $K=16$ (our default) performing best.
We hypothesize that $4$ and $8$ latents do not provide enough capacity, while $16$ strikes a good balance between expressivity and learnability.
At $K=32$, performance drops; one possible explanation is that attention becomes more diffuse over a larger latent set, making it harder for the model to learn to use each latent effectively.

\subsection{Visualizations} \label{exp:viz}
\begin{figure*}[t!]
\centering
\includegraphics[width=\linewidth]{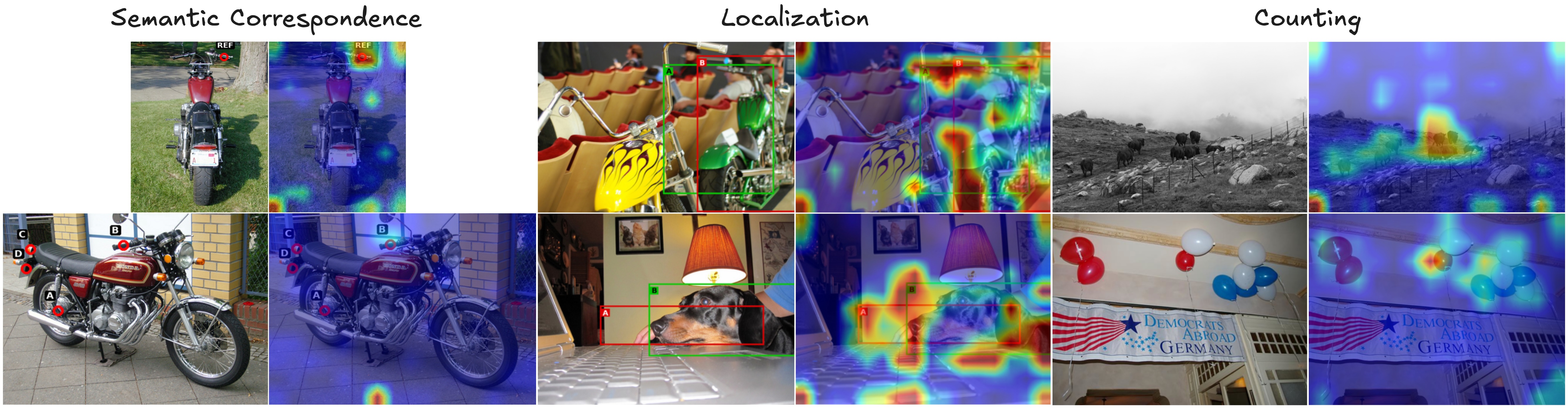}
  \caption{An illustration of latent-to-image attention maps for different tasks. The left columns show the input images, and the right columns show the attention overlays. In the Semantic Correspondence task, the model identifies the option in the second image that aligns with the REF point in the first image. In the Localization task, it selects bounding boxes that best localize the motorcycle and the dog, and in the Counting task, it counts the cows and balloons. We observe that latent-to-image attention concentrates on regions corresponding to the correct answers or the visual evidence needed to resolve each task. Although some attention sinks persist, the dominant patterns align with task-relevant regions, indicating that the latents capture meaningful visual structure without explicit supervision.}
    \label{fig:latent_viz}
\end{figure*}

\minisection{Latent Attention Visualization}
We map the latent-to-image attention maps in Figure \ref{fig:latent_viz}. Our method allows latent tokens to learn useful features across different tasks without explicit supervision. The latent tokens are able to match the handle of the motorcycle in the Semantic Correspondence task, identify the best bounding boxes of the motorcycle and the dog in the Localization task, and focus on all of the objects it needs to count in the Counting task.

\section{Conclusion}
\label{sec:conclusion}
We introduce LIVR, a method that enables LMMs to perform richer visual reasoning without requiring additional intermediate supervision. We do this by introducing latent tokens and training them with a novel visual bottlenecking approach, allowing the model to learn useful visual representations implicitly. Across nine perception-heavy tasks, LIVR consistently outperforms direct SFT in single-task training on three LMMs and improves joint multi-task training on Qwen3-VL. LIVR is also competitive with or outperforms prior reasoning methods that rely on text-based reasoning or explicit visual intermediates, while avoiding costly intermediate supervision such as helper images, bounding boxes, image crops, or chain-of-thought annotations. Through extensive experiments, we demonstrate that LIVR offers a simple, effective, and task-agnostic way to enhance visual reasoning.

\section{Limitations and Future Work}
\label{sec:Future Work}
{\looseness=-1
A limitation of LIVR is that latent tokens are less directly interpretable than text-based reasoning or hand-designed visual intermediates. Although our attention visualizations suggest that latents attend to task-relevant regions, understanding the precise information encoded by each latent remains challenging. Future work could scale LIVR to larger models, explore dynamic allocation of latent tokens based on task difficulty, and study whether latent visual reasoning can benefit more complex multi-step or interactive vision-language tasks.
}
{
    \small
    \bibliographystyle{ieeenat_fullname}
    \bibliography{main}
}
\clearpage
\appendix
\clearpage
\setcounter{section}{0}

\maketitlesupplementary

Here, we provide additional details on datasets (\Secref{supp:sec:Datasets}), training setup (\Secref{supp:sec:training}), and visualizations (\Secref{supp:sec:vis}).

\section{Datasets}
\label{supp:sec:Datasets}
For our \textbf{data-matched experiments}, we evaluate LIVR on \textbf{nine} complementary, perception-heavy tasks that together span low-level (visual correspondence, relative reflectance), mid-level (jigsaw, art style classification), and higher-level (localization, counting, visual similarity, semantic correspondence, functional correspondence) visual reasoning. Building on the BLINK benchmark and PixMo-Count, we derive evaluation splits that represent a broad and diverse testbed that stresses the model across heterogeneous task types and difficulty levels.

We provide detailed descriptions of the datasets and task setups used in our experiments.
For each of the nine perception-heavy tasks - counting, jigsaw, object localization, visual correspondence, art style, semantic correspondence, functional correspondence, relative reflectance, and visual similarity - we describe:
(i) the data sources and train/validation/test splits, and
(ii) the VQA-style prompt templates used during training and evaluation.

\subsection{Counting}
\label{supp:sec:data:counting}

\subsubsection{Data Sources and Splits}
We use the PixMo-Count dataset~\cite{pixmo}, available as
\texttt{allenai/pixmo-count} on HuggingFace, which provides an image URL, an
object label (e.g., ``people'', ``cars''), and an integer count for each
example, with official \texttt{train}, \texttt{validation}, and
\texttt{test} splits.

For training, we construct a 1,000-example subset of the PixMo-Count
\texttt{train} split by first discarding any examples whose remote image URLs
no longer resolve and then restricting to images whose ground-truth counts
lie in the range $c \in \{2,3,\dots,10\}$ and sampling examples so that these
counts are approximately uniformly represented, matching the range we evaluate
on. The PixMo authors note that the official splits may contain overlapping
images, so we perform an additional visual de-duplication step between our
\texttt{train} + \texttt{validation} images and the official PixMo-Count 
\texttt{test} images: using CLIP \cite{clip} embeddings together with perceptual hashing and SSIM-based image similarity \cite{ssim}, we flag near-duplicate pairs and remove any training/validation example whose image is a near-duplicate of a test image. From the remaining pool, we obtain 1,000 training instances.

For validation and test, we use the official PixMo-Count \texttt{validation}
and \texttt{test} splits, discarding any examples whose remote image URLs no
longer resolve; these contain 534 and 528 examples, respectively.

\subsubsection{Prompt Template}
We phrase Counting as an open-ended task, using the following prompt template:
\begin{center}
  \includegraphics[width=\linewidth]{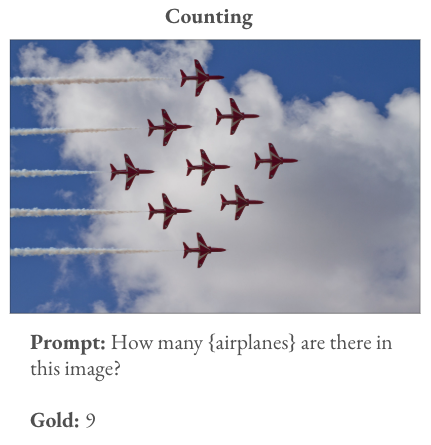}
  \label{fig:supp-count-prompt}
\end{center}

\subsection{Jigsaw}
\label{supp:sec:data:jigsaw}

\subsubsection{Data Sources and Splits}
We construct BLINK-style Jigsaw training and validation sets from COCO~\cite{coco}.
Starting from COCO2017 \texttt{train2017} and \texttt{val2017}, we
sample 1,000 and 250 images for our \texttt{train} and \texttt{val} splits,
respectively. For each image, we first sample a horizontal
canvas of fixed width $400$~px and random height in $[170, 230]$~px, then crop a
$400 \times h$ region from the original image. We partition this canvas into four
equal quadrants and treat the bottom-right quadrant as the ground-truth patch. The
input image is obtained by blacking out this quadrant, while the correct option is
the original bottom-right patch. We then sample a distractor patch of the same size
from the same COCO image, enforcing that it (i) intersects the canvas, (ii) does not
overlap the ground-truth patch, and (iii) has its center at least a fixed fraction
of the canvas size away from the ground-truth center. Finally, we randomly assign
the ground-truth patch to option A or B, and use the other as the distractor. We
ensure that each COCO image (identified by its file stem) appears in at most one of
our Jigsaw \texttt{train}/\texttt{val} splits, so there is no cross-split image
overlap within our generated data.

For our \texttt{test} split, we use the official BLINK Jigsaw \texttt{val} split (150 examples),
which is constructed from the TARA dataset~\cite{tara} rather than COCO.
Because our COCO-based Jigsaw training/validation sets and the BLINK Jigsaw
benchmark come from disjoint source datasets, we do not perform any additional
train–test de-duplication across them.

\subsubsection{Prompt Template}
We phrase Jigsaw as a two-way multiple-choice question over candidate patches, using the following prompt template:
\begin{center}
  \includegraphics[width=\linewidth]{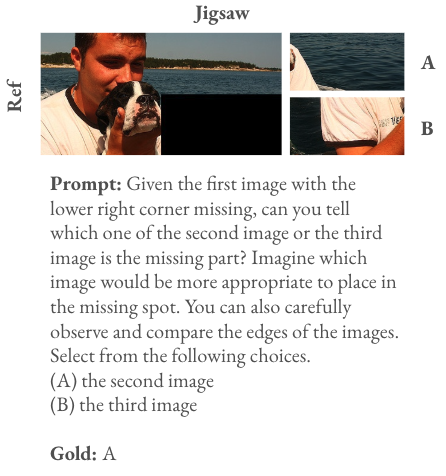}
  \label{fig:supp-jigsaw-prompt}
\end{center}

\subsection{Object Localization}
\label{supp:sec:data:localization}

\subsubsection{Data Sources and Splits}
We construct a 1,000-example training set and 250-example validation set for object localization from the COCO 2017 detection splits~\cite{coco}. Starting from the official \texttt{train2017} and \texttt{val2017} annotations, we first filter for non-crowd instances whose bounding boxes cover between 15\% and 50\% of the image area and whose segmentation masks fill at least 60\% of the box area. For each selected instance, we treat its COCO bounding box as the ``gold'' localization and generate a distractor box by jittering the four box corners until the resulting box attains an IoU in $[0.2, 0.5]$ with the gold box. We then render both boxes onto the image, label them as A and B, and phrase the task as a two-way multiple-choice question asking which box more accurately localizes the object. Because both our COCO-based data and the BLINK Object Localization benchmark are derived from the same underlying COCO imagery and overlay annotated bounding boxes, we perform an explicit visual de-duplication step between our COCO candidates (train + validation) and the BLINK Object Localization \texttt{val} images: using CLIP embeddings together with perceptual hashing and SSIM-based image similarity (on both raw and blurred grayscale images), we flag near-duplicate pairs and remove any COCO example whose image is a near-duplicate of a BLINK image. We also enforce that each COCO image (identified by its image ID) appears in at most one of our object-localization \texttt{train}/\texttt{val} splits, so there is no cross-split image overlap within our generated data. From the remaining pool of candidates, we obtain 1,000 training and 250 validation instances.

For our \texttt{test} split, we use the BLINK Object Localization \texttt{val} split (122 examples).

\subsubsection{Prompt Template}
We phrase Object Localization as a two-way multiple-choice question over candidate bounding boxes, using the following prompt template:
\begin{center}
  \includegraphics[width=\linewidth]{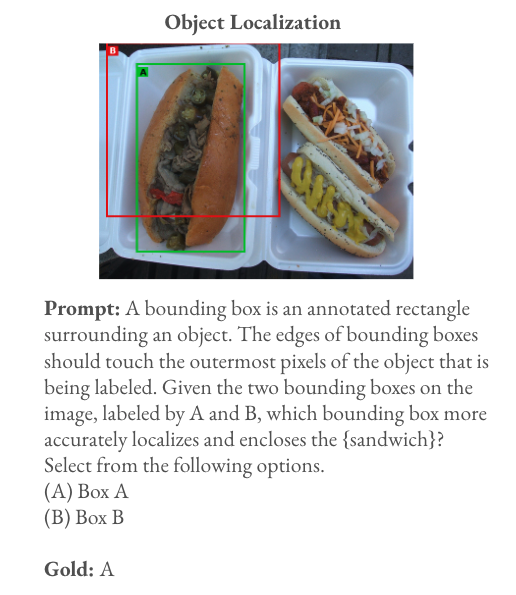}
  \label{fig:supp-localization-prompt}
\end{center}

\subsection{Visual Correspondence}
\label{supp:sec:data:viscorr}

\subsubsection{Data Sources and Splits}
We construct BLINK-style visual correspondence data from the HPatches
sequences dataset~\cite{hpatches}. We download the official HPatches
sequence release (including homographies) and treat each sequence of six
aligned images as a unit. Viewpoint and Illumination sequences are shuffled separately and then split 80/10/10 into \texttt{train}, \texttt{val}, and \texttt{test} subsets, after which we merge the viewpoint and illumination partitions for each split. This ensures that each HPatches sequence, and hence each
source/target image pair, appears in exactly one split.

Within each sequence, we consider all forward image pairs
$(i,j)$ with $1 \le i < j \le 6$ and $j - i \ge 2$, following the six-image
HPatches protocol. For a given pair, we use the provided homographies to
map points from the source image $i$ to the target image $j$. On the
source image, we repeatedly sample reference points away from the image
boundary and from one another, warp them to the target using the
$i \rightarrow j$ homography, and retain only those whose projections lie
safely inside the target image. For each valid reference point, we create
a single multiple-choice example: the source image shows the reference
point annotated as ``REF'', and the target image shows four candidate
locations (labeled A–D), one of which is the true correspondence and three
of which are distractors sampled to be far from the true location and from
each other. We randomly assign the correct correspondence to one of the
four labels, yielding BLINK-style MCQs.
Across the HPatches sequences, we generate 1,000 training,
500 validation, and 700 test examples.

Because both the BLINK Visual Correspondence benchmark and our training
data are derived from HPatches, using the official BLINK split together
with strict de-duplication would leave too little training and validation
data. Instead, we adopt the BLINK task format but evaluate on the
HPatches-based \texttt{test} split constructed above.

\subsubsection{Prompt Template}
We phrase Visual Correspondence as a four-way multiple-choice question over candidate correspondence points, using the following prompt template:
\begin{center}
  \includegraphics[width=\linewidth]{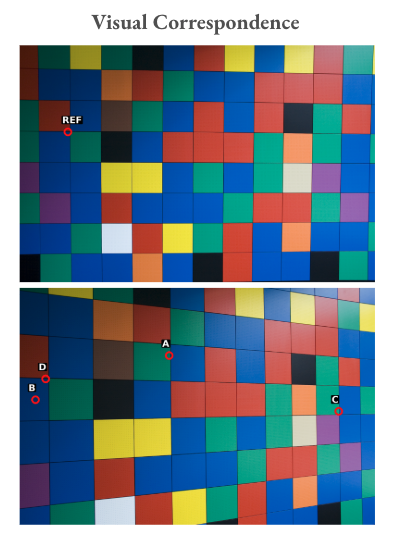}
    \includegraphics[width=\linewidth]{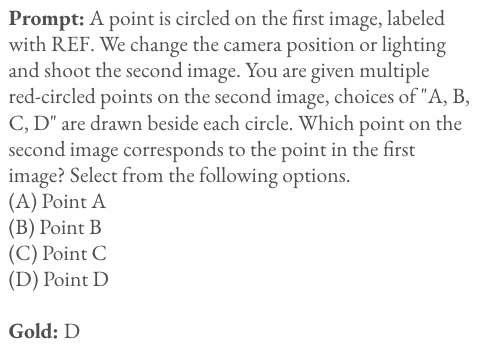}
  \label{fig:supp-viscorr-prompt}
\end{center}

\subsection{Art Style}
\label{supp:sec:data:artstyle}

\subsubsection{Data Sources and Splits}
We construct a 1,000-example training set and 250-example validation set for binary art-style classification from the ArtBench-10 dataset~\cite{artbench}. Starting from the official $256\times256$ ImageFolder variant with \texttt{train}/\texttt{test} splits, we first build a style-balanced pool of paintings from the ArtBench \texttt{train} split. Each example is then converted into a binary multiple-choice question: given a reference painting and two candidate paintings (A and B), the model must decide which candidate shares the same style as the reference. For each reference, we sample a positive candidate from the same ArtBench style (but a different image) and a negative candidate from a different style, and we randomize whether the positive candidate appears as option A or B. We ensure that each underlying ArtBench image appears in at most one of our \texttt{train}/\texttt{val} splits, so there is no cross-split image overlap within our generated data. Because ArtBench-10 also draws from large online art repositories (including WikiArt), the ArtBench and BLINK corpora can share overlapping images. To prevent train–test leakage, we perform an explicit cross-dataset de-duplication between our ArtBench-based training/validation pool (considering all three images per example: reference and both candidates) and the BLINK Art Style \texttt{val} images. Using CLIP image embeddings to retrieve high-similarity pairs, followed by perceptual hashing and SSIM-based image similarity checks, we flag near-duplicate pairs and remove any training/validation example whose images are near-duplicates of BLINK Art Style examples. From the remaining pool of candidates, we obtain 1,000 training and 250 validation examples.

For our \texttt{test} split, we use the BLINK Art Style \texttt{val} split (117 examples).

\subsubsection{Prompt Template}
We phrase Art Style as a two-way multiple-choice question over candidate images, using the following prompt template:
\begin{center}
  \includegraphics[width=\linewidth]{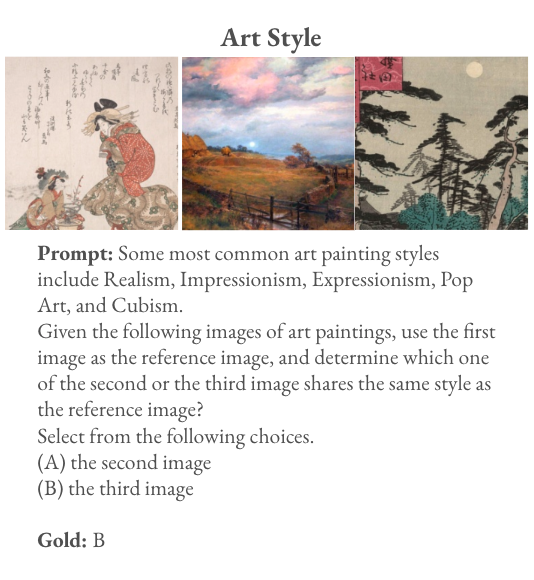}
  \label{fig:supp-art-prompt}
\end{center}

\subsection{Semantic Correspondence}
\label{supp:sec:data:semcorr}

\subsubsection{Data Sources and Splits}
For semantic correspondence, we construct a 1,000-example training set and 250-example validation set from SPair-71k~\cite{spair71k}, which provides object-category image pairs with dense keypoint annotations. We use the official training split (\texttt{trn}) and filter to pairs with at least four valid keypoint correspondences (finite, non-negative coordinates in both source and target). From this pool, we construct BLINK-style four-way MCQs: for each selected pair, we mark a single reference point on the source image at one annotated keypoint (labeled ``REF''), and on the target image we place four candidate circles (A–D), consisting of the true corresponding keypoint and three distractor keypoints sampled from the remaining annotations. All annotations are drawn directly on the original-resolution images, and we randomly assign the correct correspondence to one of the four labels. From the resulting candidates, we obtain 1{,}000 training and 250 validation examples. Because both our custom subset and BLINK's Semantic Correspondence task are derived from SPair-71k, we explicitly de-duplicate our SPair-based train/validation pool against the BLINK Semantic Correspondence \texttt{val} split.  We perform a pair-aware similarity check: for each BLINK pair, we retrieve a small set of nearest custom pairs under CLIP-based image similarity, consider both aligned and swapped orientations, and treat a pair as a duplicate only if both images pass strict perceptual-hash and SSIM criteria. Any flagged custom examples are removed. We obtain a total of 1,000 training and 250 validation instances.

For our \texttt{test} split, we use the BLINK Semantic Correspondence \texttt{val} split (139 examples) as the held-out test set, without further resampling or modification.

\subsubsection{Prompt Template}
We phrase Semantic Correspondence as a four-way multiple-choice question over candidate correspondence points, using the following prompt template:
\begin{center}
  \includegraphics[width=\linewidth]{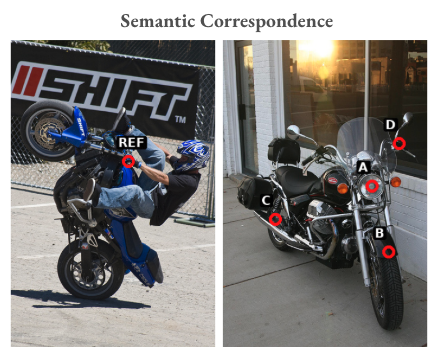}
    \includegraphics[width=\linewidth]{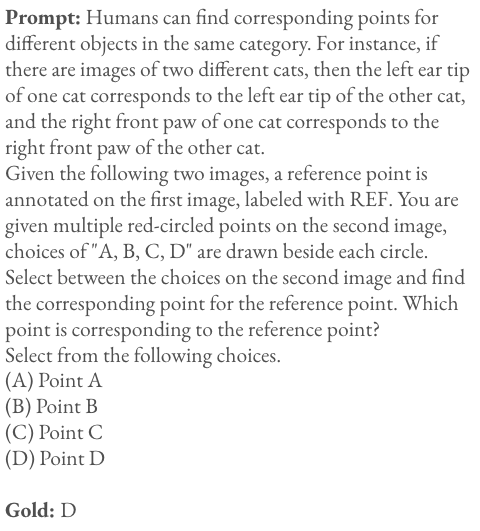}
  \label{fig:supp-semcorr-prompt}
\end{center}

\subsection{Functional Correspondence}
\label{supp:sec:data:funccorr}

\subsubsection{Data Sources and Splits}
For functional correspondence, we construct BLINK-style multiple-choice questions from the FunKPoint dataset~\cite{funkpoint}, which provides per-image action labels (e.g., ``pour'', ``scoop'') together with five normalized functional keypoints. We first form a global, disjoint 80/10/10 split of FunKPoint images into \texttt{train}, \texttt{val}, and \texttt{test}, using an action-aware balancing scheme so that the per-action image counts are approximately preserved across splits. Each underlying FunKPoint image appears in exactly one split.

Within each split and action, we then pair images that share the same action using a one-use-per-image policy: images are randomly shuffled and grouped into disjoint pairs, ensuring that no image is reused within that split and action. For a given pair, we treat one image as the left (source) and the other as the right (target). On the left image, we sample a reference keypoint index $k \in \{1,\dots,5\}$ and mark the corresponding functional keypoint with a red circle labeled ``REF''. On the right image, we annotate four candidate keypoints (A–D): the keypoint with the same index $k$ (the true correspondence) and three distractor keypoints sampled from the remaining indices $\{1,\dots,5\}\setminus\{k\}$. We randomly assign the correct correspondence to one of the four labels and phrase the task as a BLINK-style MCQ: given the action and the left ``REF'' point, select which candidate (A–D) on the right matches it. Across all actions, this procedure yields 1{,}000 training, 144 validation, and 146 test examples.

Because both our functional-correspondence data and the BLINK Functional Correspondence benchmark are derived from FunKPoint, using the official BLINK split together with strict de-duplication would leave too little training and validation data. As in our visual-correspondence setup, we therefore adopt the BLINK task format but evaluate on the FunKPoint-based \texttt{test} split constructed above, rather than on BLINK's Functional Correspondence split.

\subsubsection{Prompt Template}
We phrase Functional Correspondence as a four-way multiple-choice question over candidate correspondence points, using the following prompt template:
\begin{center}
  \includegraphics[width=\linewidth]{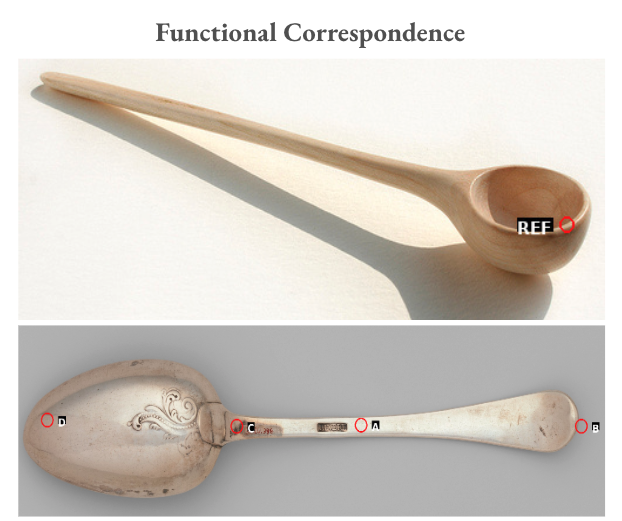}
    \includegraphics[width=\linewidth]{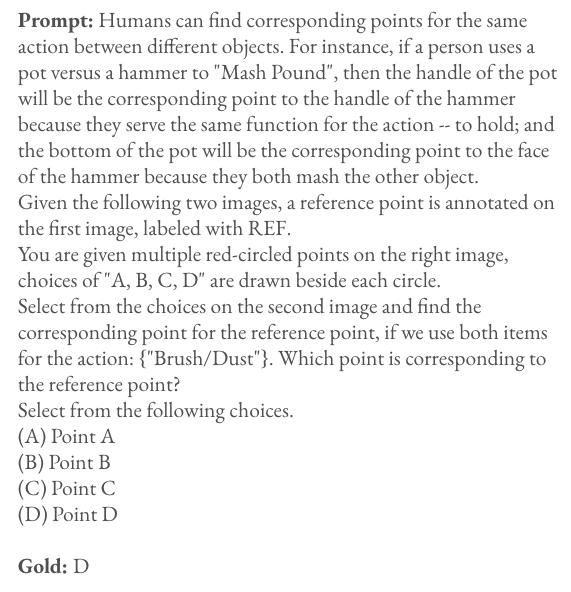}
  \label{fig:supp-func_corr-prompt}
\end{center}

\subsection{Relative Reflectance}
\label{supp:sec:data:relrefl}

\subsubsection{Data Sources and Splits}
For relative reflectance, we construct three-way multiple-choice questions from the Multi-Illumination Dataset (MID)~\cite{mid}, which provides images of indoor scenes under multiple point-light directions together with per-pixel diffuse albedo. Starting from the official MID training split, we select a subset of scenes and 25 illumination directions per scene, download sRGB images at a fixed mip level, and attach the corresponding albedo images from the released archives. For each RGB–albedo pair, we generate a single example by sampling two spatially separated points on the image, converting the albedo to linear RGB, and computing the local disk-averaged luminance at each point. Let $Y_A$ and $Y_B$ be the luminance values at the two locations; we define the relative difference
\[
\mathrm{rel} = \frac{|Y_A - Y_B|}{\max(Y_A, Y_B, 10^{-8})}.
\]
If $\mathrm{rel} \le 0.10$, we assign label (C) ``About the same''; otherwise we assign (A) ``A is darker'' or (B) ``B is darker'' according to which point has lower luminance. We control the sampling schedule so that the ``About the same'' class constitutes roughly one quarter of the data. From this generated pool, we randomly subsample 1{,}000 training and 250 validation examples.

For our \texttt{test} split, we use the BLINK Relative Reflectance \texttt{val} split (134 examples). Because our training and validation data are derived from MID while BLINK Relative Reflectance is built on IIW \cite{iiw}, no additional cross-dataset de-duplication is required.

\subsubsection{Prompt Template}
We phrase Relative Reflectance as a three-way multiple-choice question over relative surface brightness, using the following prompt template:
\begin{center}
  \includegraphics[width=\linewidth]{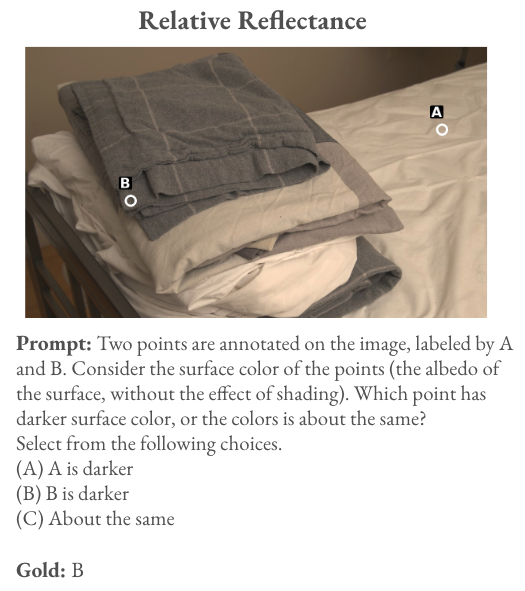}
  \label{fig:supp-reflectance-prompt}
\end{center}

\subsection{Visual Similarity}
\label{supp:sec:data:vissim}

\subsubsection{Data Sources and Splits}
For visual similarity, we use the NIGHTS dataset introduced in the DreamSim work~\cite{dreamsim} as our training and validation source. NIGHTS provides human-tested triplets consisting of a reference image and two candidates, together with votes indicating which candidate is perceptually closer to the reference. Each triplet is converted into a three-image example $(\texttt{image\_1}, \texttt{image\_2}, \texttt{image\_3})$, where \texttt{image\_1} is the reference, \texttt{image\_2} and \texttt{image\_3} are the two candidates in randomized order, and the label is ``A'' or ``B'' depending on whether the second or third image is judged more similar to the reference. Because both our NIGHTS triads and BLINK Visual Similarity are derived from the same underlying source, we enforce strict data deduplication. We deduplicate against BLINK by dropping any triad whose reference or candidate image is a near-duplicate of a BLINK Visual Similarity image, using a CLIP-based pre-filter followed by perceptual hashing and SSIM to confirm matches. After de-duplication, we retain $1{,}000$ training and $250$ validation triads from NIGHTS.

For our \texttt{test} split, we use the BLINK Visual Similarity \texttt{val} split (135 examples).

\subsubsection{Prompt Template}
We phrase Visual Similarity as a two-way multiple-choice question over candidate images, using the following prompt template:
\begin{center}
  \includegraphics[width=\linewidth]{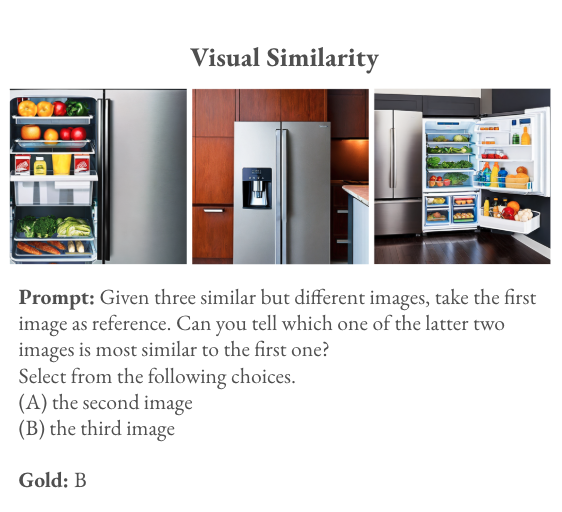}
  \label{fig:supp-vissim-prompt}
\end{center}

\section{Training Setup}
\label{supp:sec:training}

\paragraph{Single-task experiments.} Single-task experiments use the following protocol. For each of the nine perception-heavy tasks, we construct an independent training set of $1{,}000$ examples and fine-tune a separate model per task. For a given task, the Direct SFT baseline is trained for $10$ epochs on the $1{,}000$ training examples. LIVR uses a two-stage schedule with $4$ epochs of Stage~1 (masked bottleneck training) followed by $6$ epochs of Stage~2 (unmasked fine-tuning) on the same per-task training set, with $K = 16$ latent tokens. We select the checkpoint with the highest validation accuracy for reporting. 
\paragraph{Multi-task experiments.} For the multi-task setting with Qwen3-VL-4B-Instruct, we train on a combined dataset of six tasks (counting, localization, visual correspondence, semantic correspondence, functional correspondence, relative reflectance), using $1{,}000$ training examples per task ($6{,}000$ total). Direct SFT is trained for $5$ epochs, while LIVR is trained for $2$ epochs of Stage~1 and $3$ epochs of Stage~2, with $K = 16$ latent tokens. We report performance using the final checkpoint. 
\paragraph{Optimization details.} We adopt LoRA for both attention and MLP modules of the language backbone, with rank $r=16$, $\alpha=32$, and dropout $0.05$. We keep the vision encoder and projector frozen. For Direct SFT, only the LoRA parameters are trainable; for LIVR we additionally unfreeze the rows of the embedding table corresponding to the $K$ latent tokens, while all other embeddings remain frozen. We use AdamW with learning rate $1\!\times\!10^{-4}$, weight decay $0.01$, betas $(0.9, 0.999)$, and $\epsilon = 10^{-8}$. Training uses a per-device batch size of $1$ with $8$ gradient-accumulation steps, giving an effective batch size of $8$. The learning-rate schedule uses a $5\%$ linear warmup followed by cosine decay; LIVR applies a separate schedule to each stage with the same warmup and cosine decay.
\paragraph{Mirage comparison.}
For the head-to-head comparison with Mirage, we align the number of stages and epochs but keep each method’s own two-stage objective. For LIVR, we set $K = 4$ latent tokens and train for $10$ epochs in Stage~1 (LIVR’s masked bottleneck) and $10$ epochs in Stage~2 (LIVR’s unmasked fine-tuning). For Mirage, we likewise use $K = 4$ latent tokens, training for $10$ epochs with Mirage’s Stage~1 objective and $10$ epochs with Mirage’s Stage~2 objective. In this comparison we use full fine-tuning of the language backbone (while keeping the vision encoder and projector frozen), with AdamW (learning rate $1\!\times\!10^{-5}$, weight decay $0.01$, betas $(0.9, 0.999)$, $\epsilon = 10^{-8}$) and the same effective batch size of $8$ (per-device batch size $1$ with $8$ gradient-accumulation steps). The learning-rate schedule again uses a $5\%$ linear warmup followed by cosine decay for each stage.

\paragraph{Compute resources.}
All experiments are run with a fixed random seed of $42$ on a single node equipped with $8$ NVIDIA RTX 6000 Ada GPUs, using single-GPU training for each run.

\section{Additional Visualizations}
\label{supp:sec:vis}

%

 \begin{figure}[htbp!]
   \includegraphics[width=\columnwidth]{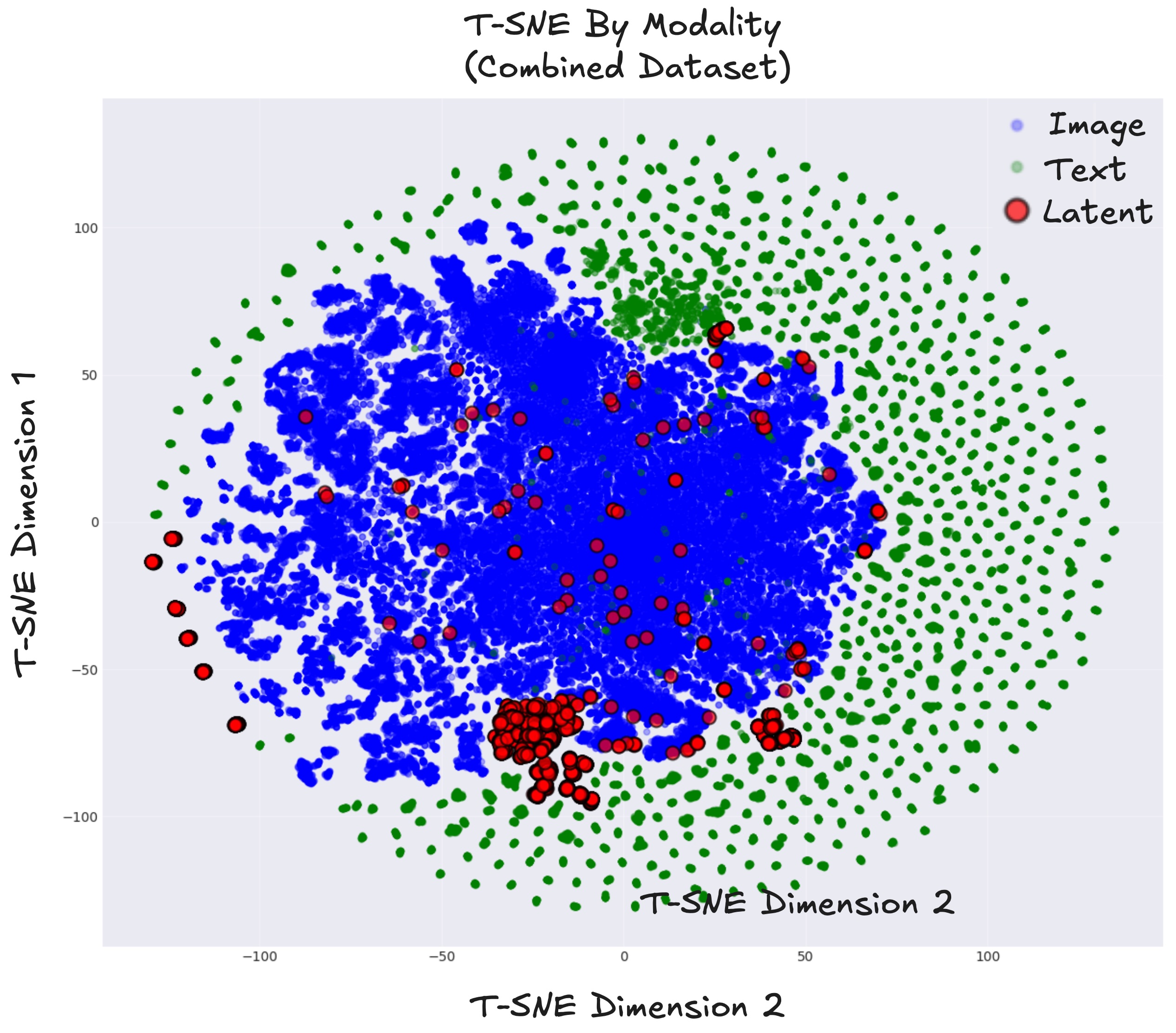}
   \caption{t-SNE Visualization of Different Tokens}
   \label{fig:supp-tsne}
   \vspace{-10pt}
 \end{figure}

 \minisection{Latent Token t-SNE}
 To probe what our latent tokens represent, we visualize token hidden states using t-SNE. We first train Qwen3-VL-4B-Instruct in the multi-task setting on the six-task mixture used in Sec.~\ref{sec:multitask}. We then extract the final-layer hidden states for all token positions (latent, image, and text) from the first 50 evaluation examples of each of the six tasks, and embed them with t-SNE (Fig.~\ref{fig:supp-tsne}; red: latent, blue: image, green: text). Latent tokens largely occupy the same region as image tokens in the t-SNE projection, suggesting that many latent representations align with the model’s visual feature space. At the same time, a compact cluster of latent tokens forms a distinct region, suggesting that some latents learn specialized representations not fully captured by the image-token manifold.

\begin{figure*}[t!]
\centering
  \includegraphics[width=\linewidth]{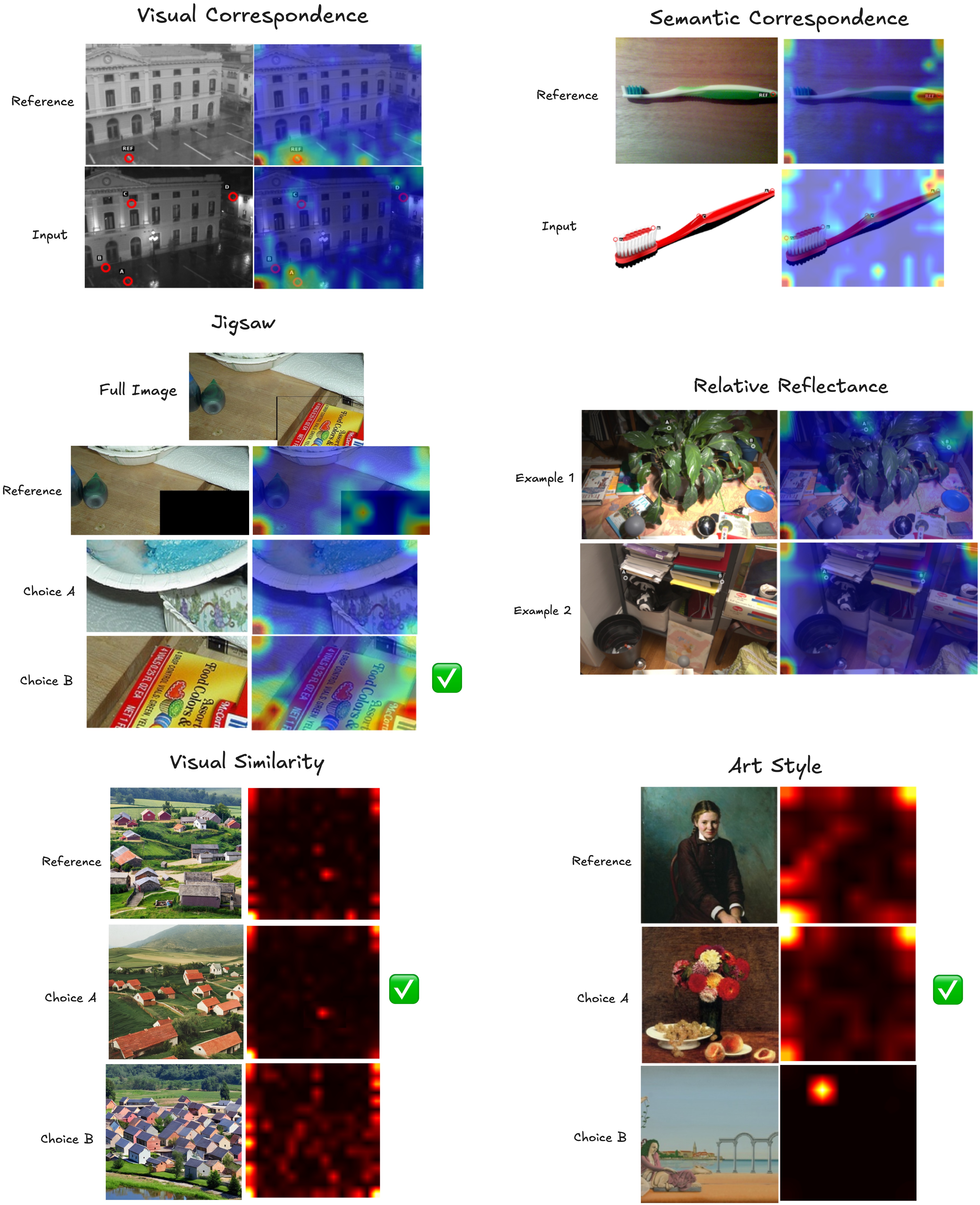}
  \caption{Additional visualizations of latent token to image attention.}
  \label{fig:supp-viz-add}
\end{figure*}
\subsection{Additional Latent-to-Image Attention Visualizations}
We show in Figure \ref{fig:supp-viz-add} some additional examples of latent-to-image attentions for Qwen-3-VL-4B. For the Visual Similarity and Art Style tasks, we display the raw attention maps instead. For question types where images serve as the choices (Jigsaw, Visual Similarity, and Art Style), we display a check mark symbol next to the correct answer. We can see that in the Visual Correspondence example, the latent tokens focus correctly on the parking line that is indicated by the REF point. In the Semantic Correspondence task, the model focuses correctly on the toothbrush handle. In the Jigsaw task, the latent tokens actually focus on the diagonal ledge of the table in the masked out reference image, and finds this feature in Choice B, which is the correct answer. In Relative Reflectance, the latent tokens focus on the 2 points that it needs to compare. Finally, in the Visual Similarity and the Art Style tasks, the attention maps for the latent tokens of the correct answers are much more similar to attention maps of the reference images. For these tasks, it may be hard for humans to define explicit representations, but our approach allows latent tokens to learn these useful representations implicitly.

\end{document}